**Emulating Complex Synapses Using Interlinked Proton Conductors**

Lifu Zhang[1], Ji-An Li[2], Yang Hu[1], Jie Jiang[1], Rongjie Lai[3,*], Marcus K. Benna[4,*], Jian Shi[1,5*]

[1]Department of Materials Science and Engineering, Rensselaer Polytechnic Institute, Troy, NY 12180, United States
[2]Neurosciences Graduate Program, University of California San Diego, La Jolla, CA 92093, United States
[3]Department of Mathematics, Purdue University, West Lafayette, IN 47907, United States
[4]Department of Neurobiology, School of Biological Sciences, University of California at San Diego, La Jolla, CA 92093, United States
[5]Department of Physics, Applied Physics, and Astronomy, Rensselaer Polytechnic Institute, Troy, NY 12180, United States
[*]Corresponding authors: R.L.: lairj@purdue.edu, M.K.B.: mbenna@ucsd.edu, J.S.: shij4@rpi.edu

## Abstract

In terms of energy efficiency and computational speed, neuromorphic electronics based on non-volatile memory devices is expected to be one of most promising hardware candidates for future artificial intelligence (AI). However, catastrophic forgetting, networks rapidly overwriting previously learned weights when learning new tasks, remains as a pivotal hurdle in either digital or analog AI chips for unleashing the true power of brain-like computing. To address catastrophic forgetting in the context of online memory storage, a complex synapse model (the Benna-Fusi model) has been proposed recently[1], whose synaptic weight and internal variables evolve following a diffusion dynamics. In this work, by designing a proton transistor with a series of charge-diffusion-controlled storage components, we have experimentally realized the Benna-Fusi artificial complex synapse. The memory consolidation from coupled storage components is revealed by both numerical simulations and experimental observations. Different memory timescales for the complex synapse are engineered by the diffusion length of charge carriers, the capacity and number of coupled storage components. The advantage of the demonstrated complex synapse in both memory capacity and memory consolidation is revealed by neural network simulations of face familiarity detection. Our experimental realization of the complex synapse suggests a promising approach to enhance memory capacity and to enable continual learning.

## Introduction

Conventional artificial neural networks have had tremendous success in many problems in science and engineering[2-10]. However, biological learning and memory systems are far more energy efficient than the state-of-the-art artificial intelligence computing systems that have been demonstrated (e.g. 20 W for human brains versus 170 kW for AlphaGo)[11]. Despite significant efforts and progress made on brain-inspired computing in the past decades, existing neuromorphic hardware struggles in competing with traditional digital and analog processors in terms of efficiency for many artificial intelligence tasks[12], let alone with human brains. A fundamental reason leading to such a gap is that most brain-inspired hardware only harnesses very limited features of their biological counterparts[1, 13, 14]. One typical example is the emulation of synaptic learning and memory[1, 15-17].

Current mainstream approaches in developing and engineering artificial synapses focus on single variable designs that are largely inspired by classical rate networks such as the Hopfield model, whose simple synapses are described entirely by their efficacy or weight. Recent progress in theoretical neuroscience shows that complex synapses with multiple variables may offer far superior memory capacity and performance in (unsupervised) online learning scenarios, especially if nonlinearities limit their dynamical ranges[1, 18]. Further, the metaplastic behaviors of complex synapses with hidden variables may offer superior performance for continual learning of multiple tasks in sequence[19-21], suggesting rich opportunities for using complex synapse models in deep learning circuits to ameliorate issues of catastrophic forgetting, namely, the rapid forgetting of previously learned weights when training a network on a new task (Fig. 1a and b).

Fig. 1 illustrates a possible scenario for continual learning. The learning trajectory (dashed line in Fig. 1c) of simple synapses is independent from past experience (except for its starting point, since there is no place for storing the memory of previous tasks other than in the synaptic weights). This can be understood from the analogy with a single beaker of size $C_1$ (Fig. 1e) whose liquid level $u_1$ denotes the synaptic weight $w_{ij}$. The learning trajectory (dashed line in Fig. 1d) of complex synapses, however, is regulated or constrained by the past experience. In Fig. 1d, we show that the learning path of complex synapses for task 2 lies in the colored region illustrating the subspace of the network weights suitable for performing task 1. In this case, the final learned parameters sit in the overlapping region of tasks 1 and 2, indicating successful continual learning. This process can be illustrated by a coupled multiple-beaker system in which the change of liquid height in the $C_1$ beaker for learning task 2 is influenced by the memory (the liquid height $u_2$) for task 1 (stored in the $C_2$ beaker) flowing back to $C_1$. This does not occur for simple synapses: after learning, the parameters sit outside the previous task's preferred region (e.g. outside the region of task 1 in Fig. 1c) – this implies catastrophic forgetting.

The complex synapse model proposed by Benna and Fusi in 2016 (also known as Benna-Fusi model[1]), combines multiple dynamical processes that initially store memories in fast variables and then progressively transfer them to slower variables. In this model, the online-learning memory capacity scales almost linearly with the number of synapses, which is a substantial improvement over the square root scaling of previous single-variable models with low-precision weights. In the complex synapse (right panel of Fig. 1e) represented by a series of connected beakers (of cross-sectional area $C_k$), the liquid level of the leftmost beaker ($u_1$) corresponds to the synaptic weight, and other connected beakers store the memory trace at longer timescales.

Inspired by these theoretical advances, in this work, we demonstrate a hardware realization of an artificial complex synapse based on the Benna-Fusi model and reveal its superior memory capacity during online learning for real-world applications such as face familiarity detection. In detail, we choose the electrochemically active organic material poly(3,4-ethylenedioxythiophene): polystyrene sulfonate (PEDOT:PSS) to design the complex synapse, whose electronic conductivity is determined by the proton concentration[22, 23]. By designing a polymer-based redox transistor together with a series of charge-diffusion-controlled storage components where Nafion is utilized as the diffusion medium, a complex synapse can be realized (Fig. 1f). We engineer different timescales for the synaptic variables by controlling both the diffusion length ($g_{k,k+1}$ as in Fig. 1e) and the size of storage components ($C_k$ as in Fig. 1e) to demonstrate memory consolidation. Our observation of complex synapse-like memory dynamics sheds some light on the design of AI hardware to unveil the full power of brain computing.

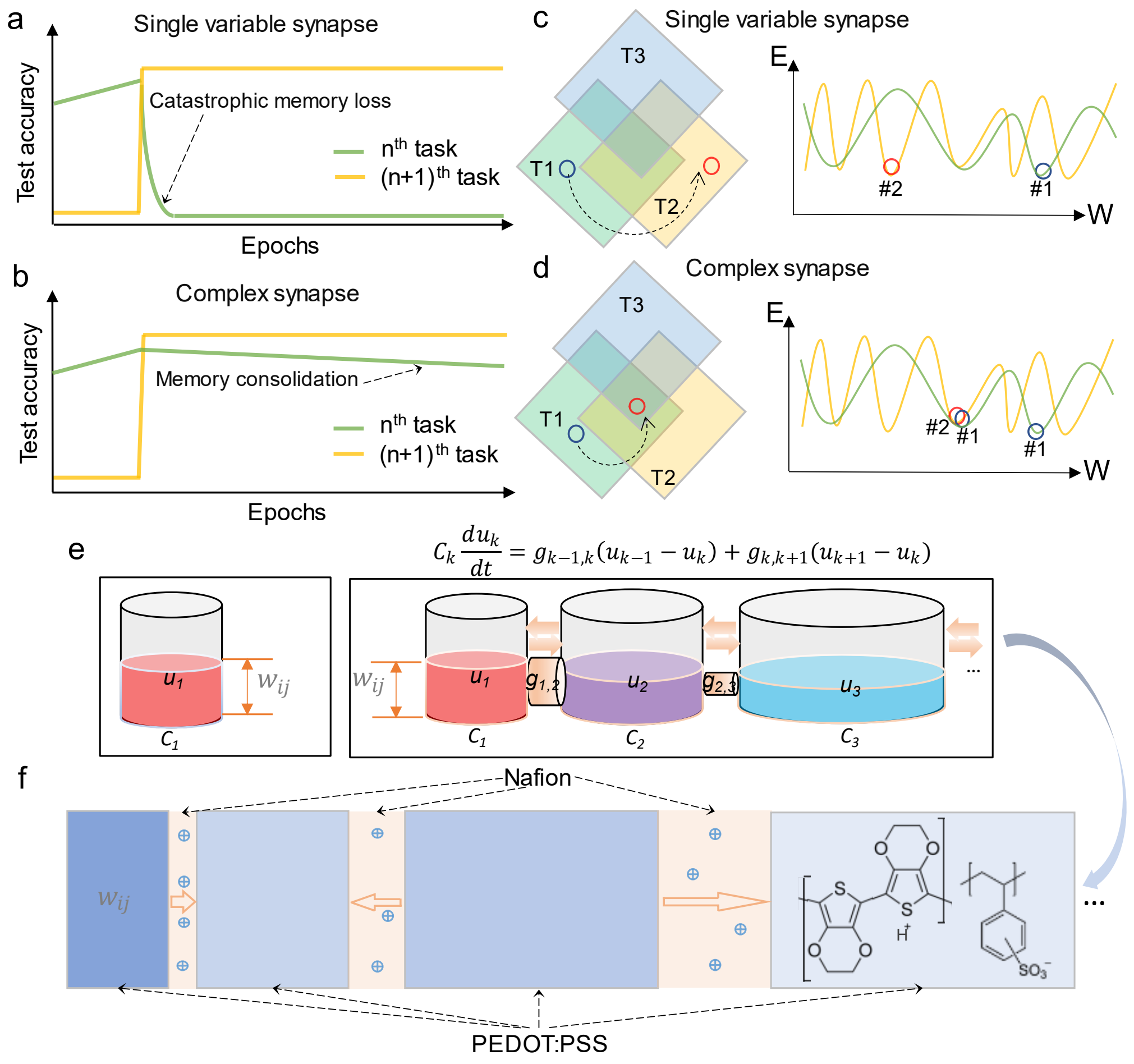

*Fig. 1* ***Complex synapse and synaptic memory consolidation in continual learning of multiple tasks. a****, Catastrophic forgetting for a network with single variable synapses when training a new task.* ***b****, Alleviation of catastrophic forgetting for a network with complex synapses due to memory consolidation.* ***c****, Trajectory of synaptic weight sets in the weight space during multi-task learning for a network with single-variable synapses (left panel) and the corresponding energy levels of the synaptic weight sets during learning (right panel). Weight space for different tasks: light green (task 1), light yellow (task 2) and light blue (task 3). The weight set after updates (#1→#2) reach a local minimum for task 2 but is nearly a local maximum for task 1.* ***d****, Trajectory of synaptic weight sets in the weight space during multi-task learning for a network with complex synapses (left panel) and the corresponding energy levels of the synaptic weight sets during learning (right panel). The weight set updating (#1→#2) is regulated by the past memory in complex synapses.* ***e****, Schematic of Benna-Fusi complex synapse (right panel) and a single variable synapse (left panel). The $u_k$ variable measures the liquid level of $k_{th}$ beaker, $C_k$ scales with the size (area) of the $k_{th}$ beaker, and the coupling coefficient $g_{k,k+1}$ scales with the cross-section of the connecting tube between the $k_{th}$ and $(k+1)_{th}$ beakers. The liquid level in the first beaker (red) represents the synaptic strength $w_{ij}$. A differential equation on the top right gives the weight update dynamics of a complex synapse.* ***f****, Design of a hardware device for artificial complex synapse based on the Benna-Fusi model. PEDOT:PSS is utilized as storage components ($C_k$) and Nafion is used as coupling layers where the coupling coefficient $g_{k,k+1}$ can be controlled by the lateral lengths of Nafion film. Charges (protons and electrons) can diffuse between neighboring storage components within the media of Nafion. The degree of blue for each storage component represents different concentrations of proton in PEDOT:PSS, by which the electronic conductivity of PEDOT:PSS is determined.*

## Results

The complex synapse is characterized by a series of storage components $C_k$ and coupling layers between neighboring storage components $g_{k,k+1}$. Each storage component stores the memory species, proton in PEDOT:PSS, and the Nafion coupling layers regulate the diffusion of proton between neighboring storage components. The instantaneous synaptic weight is reflected by the amount of the memory species (proton concentration) in $C_1$. The time-dependent synaptic weight of $C_1$ is influenced by all other $C_k$ components at different timescales. Single-variable synapses typically require only a single storage component, and no coupling layer is needed. In our proposed synapse, its hierarchical structure requires reading and writing procedures to be operated through different sets of electrical terminals, as shown in the schematic of Fig. 2a. Instantaneously removing/adding memory species from/to the complex synapse (through $C_1$) can be achieved by using a reservoir component R with a gate voltage $V_G$ through gate electrode G, which is an electrochemical redox process[24]. The reading operation is conducted by measuring the electrical conductivity through a source electrode S and drain electrode D, since there is a nearly linear dependence between electrical resistivity and proton concentration of PEDOT:PSS[25, 26].

We fabricate such complex synapse devices with an optimized semiconductor processing procedure. Briefly, metal interconnects for the redox transistor were first patterned using a standard lift-off process on a silicon wafer with 300 nm thermal oxide, and the wafer was then coated with ~3 μm of Parylene-C as the shadow layer. Corresponding photolithographic patterning was performed on the coated wafer and thoroughly etched to define the transistor channel/gate and areas for other storage components. PEDOT:PSS was spin-coated on the wafer and well-defined storage components were formed by finally exfoliating the Parylene-C shadow layer (more information can be found in the Methods). A complex synapse device with two storage components is displayed in the bottom panel of Fig. 2a. Nafion gel-like electrolyte was prepared from solution by drop-cast method.

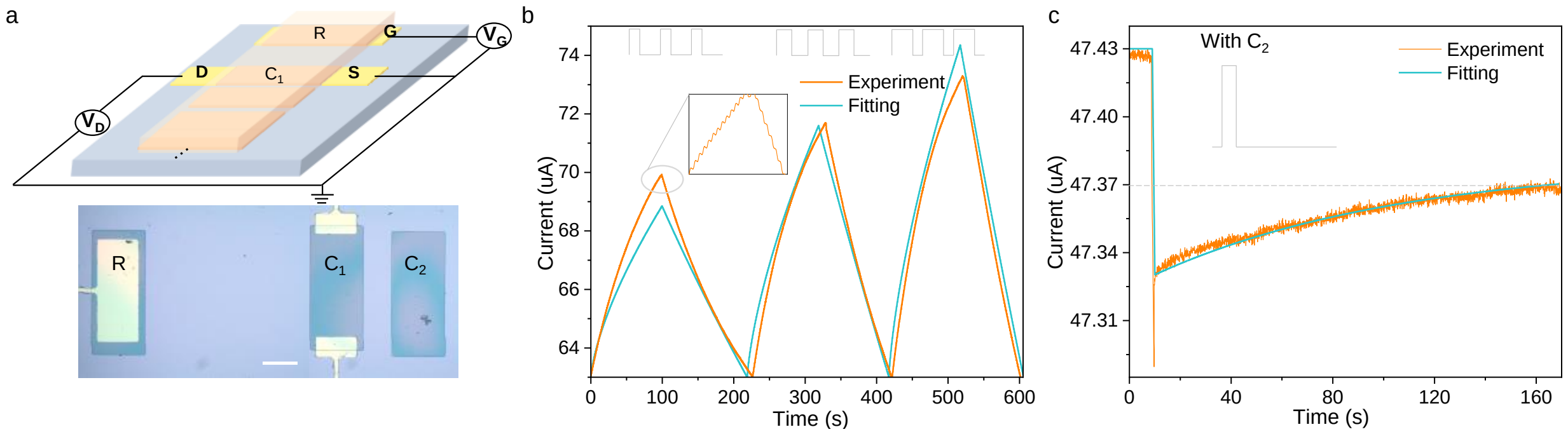

*Fig. 2* ***Characterization of PEDOT:PSS-based synapses. a****, Schematic of the PEDOT:PSS-based synapse (top panel) and an optical image of a real device (bottom panel). Scale bar: 200 μm.* ***b****, Synaptic potentiation and depression of a single-variable synapse (only $C_1$) with writing pulses of -/+1 V but of different "on" times within each training pulse (within one training pulse, on state means writing voltage is on; off state means writing voltage is zero). "on"-state fraction (defined by the period of the on state over the whole period within one pulse cycle) from left to right: 33%, 50%, 67%. Pulse frequency: 1 Hz. Pulse number for each depression process is the same: 100.* ***c****, The weight change of a complex synapse (with additional storage component $C_2$) as a function of time after a large writing pulse (+3 V, 1 s "on" time). The coupling coefficient $g_{12}$ can be extracted from the simulated curve. A voltage of 0.01 V is adopted for all the read operations.*

To evaluate the synaptic plasticity of the PEDOT:PSS-based redox transistor, we first investigate the update of synaptic states for a single-variable synapse systematically with different writing schemes (Fig. S1). Fig. S1a-b show the synaptic potentiation and depression of a single-variable synapse (only $C_1$) with writing pulses of -/+1 V. Each writing pulse includes a period of the writing voltage being on ($t_{on}$) and a period of the writing voltage being zero ($t_{off}$). As we can see, the total change of synaptic weight for the 0.5 Hz/25 pulses condition is close to that for the 1 Hz/50 pulses condition, which is as expected. Numerical simulations under these two conditions can largely reproduce our experimental results (Fig. S1d and e, with the amplitude of *s* fixed to be the same, where *s* represents the weight change per unit time due to the imposed writing pulse. See more descriptions in Supplementary Discussions). For the simulation method and details, more information can be found in Supplementary Materials.

Fig. S1c shows the synaptic potentiation and depression of a single-variable synapse (only $C_1$) with writing pulses of different voltages. In the numerical simulations (Fig. S1e-f), *s* for pulse strength of ±2 V (Fig. S1f) is set as twice of that for pulse strength of ±1 V (Fig. S1e). Comparing the simulated data with the experimental results (Fig. S1b and e, Fig. S1c and f), their consistency further indicates effective controllability of weight updates in our devices via gate voltage pulses. The repeated weights patterns controlled by writing voltages with different "on" state fractions (ratio of the "on" state time over the full period of one training pulse) and amplitudes shown in Fig. S1 also verify that multiple synaptic states can be reliably achieved in our PEDOT:PSS-based redox transistors. Another feature we can observe from these curves is that the potentiation part is almost symmetric to the depression part. However, we find this feature disappears when the updated weight range becomes large (Fig. S2, for example). In Fig. S2, there are 80 writing pulses for potentiation of a single-variable $C_1$ while it takes more than 100 pulses for depression to its original state. *s* becomes smaller and smaller as the weight increases to a high level for the potentiation process. On the contrary, the corresponding *s* for the depression process is large when the weight value is high, and it becomes smaller and smaller as the weight decreases to a lower level. This nonlinearity together with asymmetric potentiation/depression for large weight dynamic ranges has been observed in different material/device systems, considering their complex kinetics in both forward and backward directions[27-34].

Fig. 2b displays the potentiation and depression of a single-variable synapse with writing pulses of -/+1 V but of different "on"-state fractions (33%, 50%, 67% from left to right). The total weight change for potentiation (80 writing pulses) increases as the "on"-state fractions increases, which is as expected. Furthermore, the total number of writing pulses for depression to its original weight level is different for different "on"-state fractions. For the highest "on"-state fraction (67%), the single-variable synapse requires less writing pulses to depress to its original synaptic level. This is plausible since higher weight levels can be reached with a higher "on"-state fraction, and for a depression process *s* becomes much larger when the weight level is high (see inset of Fig. 2b). The simulated result for a single-variable synapse under such working conditions is also shown in Fig. 2b. Similar trends can be extracted for the simulated curves.

To reveal the contributions from additional storage components ($C_k$, $k > 1$) of a complex synapse, we use a single voltage pulse to stimulate our designed complex synapse and make a comparison with the single-variable synapse, as shown in Fig. 2c and Fig. S3. Fig. S3 displays the weight change of a single-variable synapse after a single voltage pulse of +3 V (1 s "on" time). There is almost no change for the updated weight after a long period. However, the observation is totally different for our complex synapse with additional $C_2$ component. There is around 40% recovery of the weight change within 160 s after the single writing pulse (Fig. 2c). The synapse becomes resilient to weight changes with the addition of coupled storage components. Based on the simulation for a complex synapse with additional $C_2$ under this working condition (Fig. 2c, cyan curve), the coupling coefficient between $C_1$ and $C_2$ can be determined to be $g_{12}/C_1 = 2^{-7.5}$/second. We can see that coupled storage components in our complex synapse give the synapse a tendency to recover towards its original state, thereby implementing a useful compromise between storing new information and retaining previously stored memories.

A systematic study for the effects of additional storage components ($C_k$, k > 1) in PEDOT:PSS-based complex synapse is displayed in Fig. 3. Fig. 3a shows PEDOT:PSS-based a complex synapse array fabricated on a two-inch wafer with traditional semiconductor-processing techniques (more information about the fabrication procedure can be found in Fig. S5). Fig. 3b shows the synaptic potentiation-depression cycles of a simple synapse (single variable) with writing pulse of different “on” state fractions. The total pulse number for each potentiation process is the same, while the depression pulse numbers required to recover to its original state can be different, due to the asymmetry of our devices as mentioned in Fig. 2b and Fig. S2. As we can see, the total weight change for potentiation (80 writing pulses) is largest for the pulse “on” state fraction of 67% and it reduces as the “on” state fraction decreases. For the depression process, on the contrary, a longer period is needed when the pulse “on” state fraction becomes smaller.

A same writing procedure has been conducted on this synapse but with the additional $C_2$ storage component connected (Fig. 3c). Compared with the result in Fig. 3b, the total weight change for potentiation using the same pulses becomes smaller for the complex synapse. The depression time required to go back to its original state of the complex synapse for each depression process also reduces, correspondingly. This implies that the synapse becomes more resilient with the coupled storage component $C_2$. An illustration of such a “regularizing effect” from additionally coupled storage components in a complex synapse during a potentiation-depression cycle is shown in Fig. 3d. The updated changes of weight for both simple and complex synapse during this whole potentiation process with 80 writing pulse of different “on” state fractions are summarized in Fig. 3e.

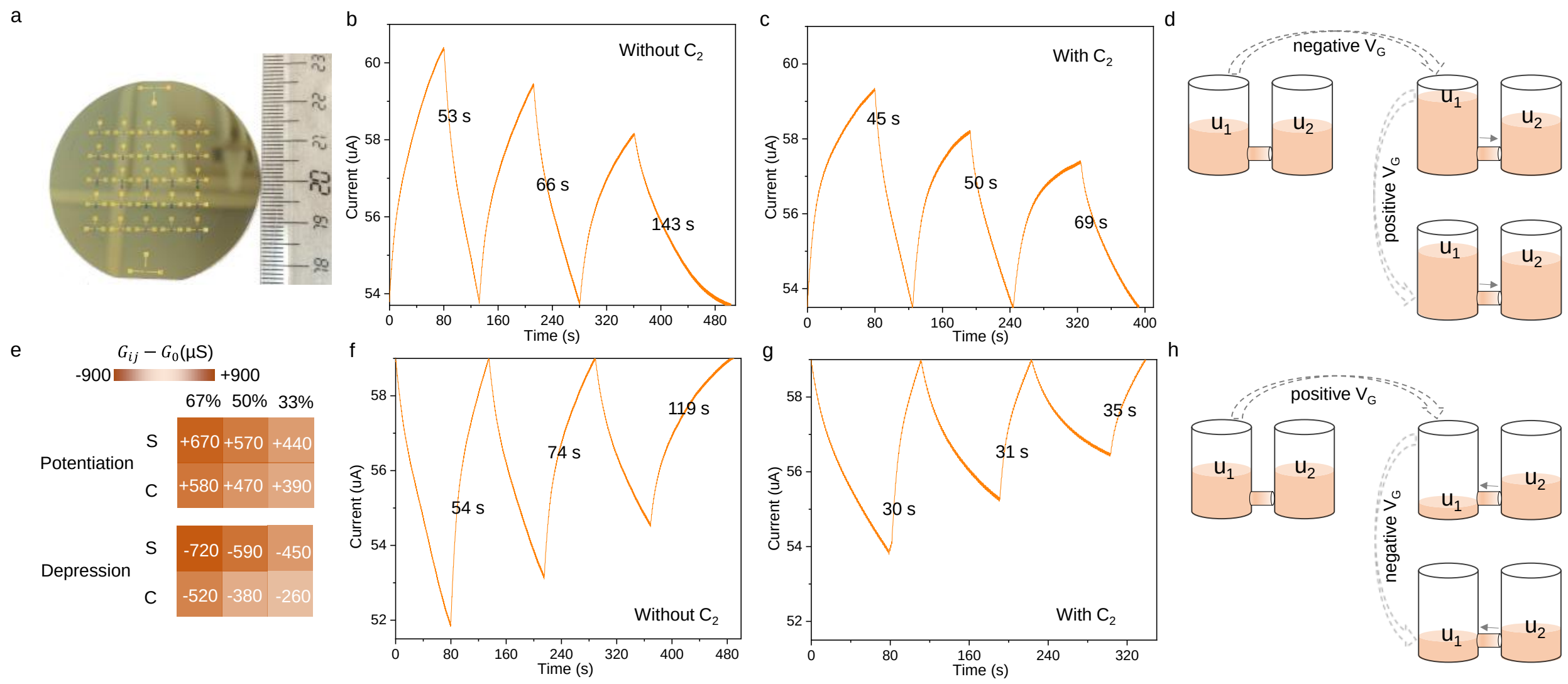

*Fig. 3* ***Effect of coupled storage components ($C_k$, k > 1) in PEDOT:PSS-based complex synapses. a****, Photograph of PEDOT:PSS-based complex synapse array fabricated on a wafer.* ***b****, Synaptic potentiation and depression of a simple synapse (only $C_1$) with writing pulses of -/+1 V but of “on” state fractions. “on” state fractions from left to right: 67%, 50%, 33%. Pulse frequency: 1 Hz. Pulse number for each potentiation process is the same: 80.* ***c****, Synaptic potentiation and depression of the same synapse in* ***b*** *but with the additional $C_2$ connected. The writing procedure is the same as that in* ***b****.* ***d****, Illustration of the effect of additional storage components in a complex synapse during a potentiation-depression cycle.* ***e****, Updated changes of weight during a whole potentiation or depression process of 80 pulses (-/+1 V). “on” state fraction: 67%, 50%, 33%. Pulse frequency: 1 Hz. S and C indicate simple synapse and complex synapse (with coupled $C_2$), respectively.* ***f****, Synaptic depression and potentiation of a simple synapse (only $C_1$) with writing pulses of +/-1 V but of different “on” state fractions. “on” state fractions from left to right; 67%, 50%, 33%. Pulse frequency: 1 Hz. Pulse number for each depression process is the same: 80.* ***g****, Synaptic depression and potentiation of the same synapse in* ***f*** *but with the additional $C_2$ connected. The writing procedure is the same as that in* ***f****.* ***h****, Illustration of the effect of additional storage components in a complex synapse during a depression-potentiation cycle. A voltage of 0.01 V is adopted for all the read operations.*

Fig. 3f displays synaptic depression-potentiation cycles of a simple synapse with writing pulses of different "on" state fractions. Similar results can be found under this reversed writing procedure (compared to results in Fig. 3b). The total weight change for depression (80 writing pulses) is largest for the pulse "on" state fraction of 67% and it reduces as the "on" state fraction decreases. On the other hand, a longer period is needed for the potentiation process to recover to its original state when the pulse "on" state fraction becomes smaller. The same writing procedure has been conducted on this synapse again but with the additional $C_2$ storage component connected (Fig. 3g). As we can see, the total weight change for depression of same pulses becomes smaller for the complex synapse (summarized in Fig. 3e, bottom panel) and the potentiation time required to recover to its original state for each potentiation process also reduces, correspondingly. An illustration of the "regularizing effect" from additionally coupled storage components in a complex synapse during a depression-potentiation cycle is displayed in Fig. 3h. Simulations for both simple and complex (with extra $C_2$ storage component) synapses under these two reversed writing procedures (potentiation-depression and depression-potentiation) are shown in Fig. S4. They are consistent in trends with the experimental results of Fig. 3b-c and Fig. 3f-g.

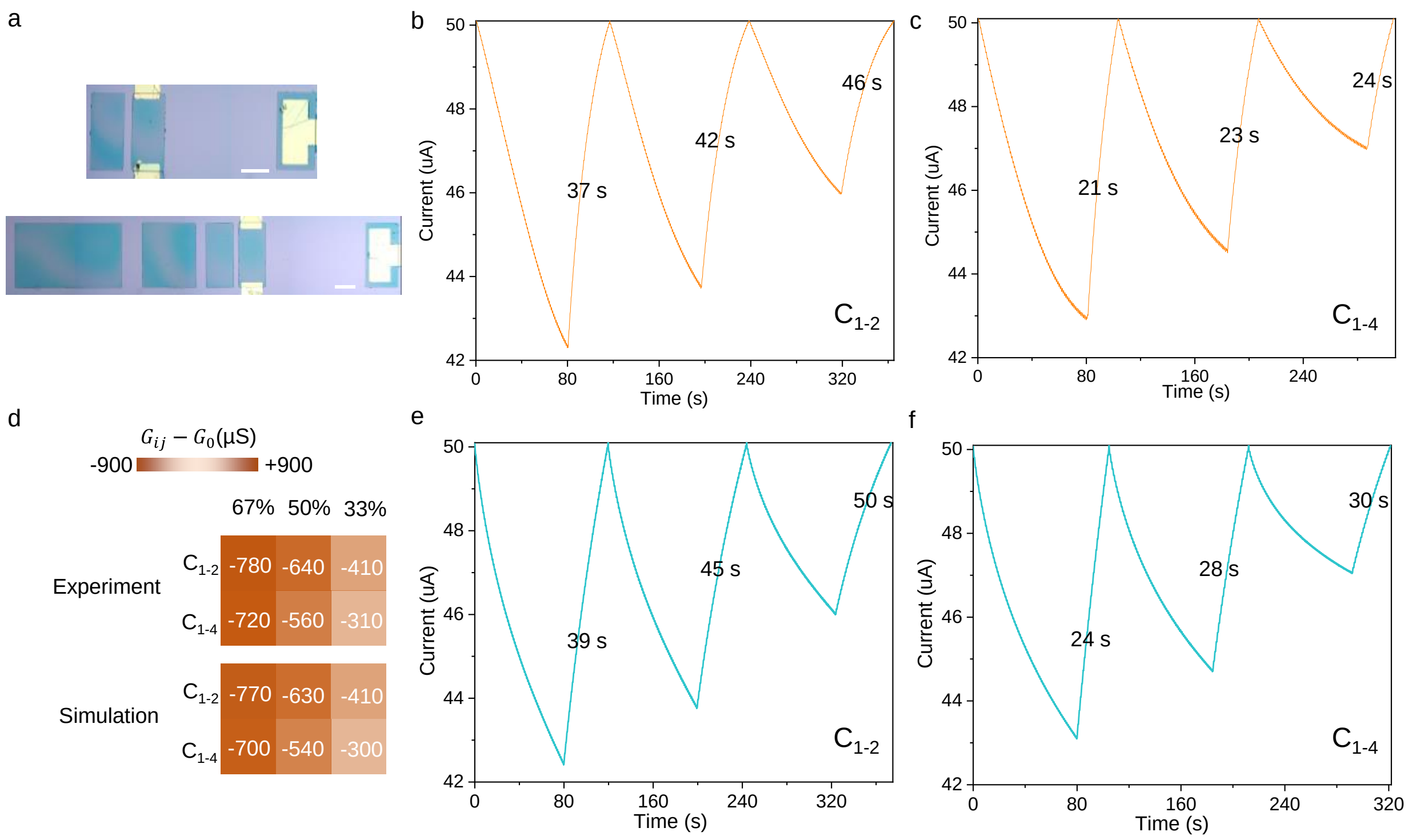

***Fig. 4 Effect of different coupled components ($C_k$, $k > 1$) in PEDOT:PSS-based complex synapses. a**, Optical images of two devices used in **b** and **c**. Scale bar: 200 μm. **b**, Synaptic depression and potentiation of a complex synapse (with additional $C_2$) with writing pulses of +/-1 V but of different "on" state fractions. "on" state fractions from left to right; 67%, 50%, 33%. Pulse frequency: 1 Hz. Pulse number for each depression process is the same: 80. **c**, Synaptic depression and potentiation of a complex synapse (with additional $C_2$, $C_3$ and $C_4$) with writing pulses of +/-1 V but of different "on" state fractions. "on" state fractions from left to right; 67%, 50%, 33%. Pulse frequency: 1 Hz. Pulse number for each depression process is the same: 80. **d**, Updated changes of weight during a whole depression process of 80 pulses (+1 V) for complex synapses with different coupled components. **e**, Simulation of synaptic depression and potentiation of a complex synapse (with additional $C_2$) with writing pulses of +/-1 V but of different "on" state fractions. "on" state fractions from left to right; 67%, 50%, 33%. Pulse frequency: 1 Hz. Pulse number for each continuous depression process is the same: 80. **f**, Simulation of synaptic depression and potentiation of a complex synapse (with additional $C_2$, $C_3$ and $C_4$). The writing procedure is the same as that in **e**. Coupling coefficients of $g_{12}: g_{23}: g_{34} \cong 2^{-6}: 2^{-7}: 2^{-8}$ for these complex synapses can be extracted from the simulated results. A voltage of 0.01 V is adopted for all the read operations.*

We then study the effect of the number and size of the coupled storage components and the related coupling coefficients on its memory timescale. Fig. 4 displays the synaptic depression and potentiation of two complex synapses containing different numbers of coupled storage components ($C_2$ and $C_2$-$C_4$) with writing pulses of different "on" state fractions. As we can see, the total weight change during depression with the same pulses (80 pulses) is less for the complex synapse with more coupled storage components, comparing cycles of the same "on" state fractions in Fig. 4b and c. The potentiation time required to recover to the original state of the synapse is also less for the complex synapse with more coupled storage components. These observations clearly support our hypothesis that different memory timescales can be achieved by adjusting variables of the coupled storage components. Based on our simulation (Fig. 4e and f), the coupling coefficients can be determined to be $g_{12}/C_1 : g_{23}/C_1 : g_{34}/C_1 \cong 2^{-6} : 2^{-7} : 2^{-8}$ for the measured devices as shown in Fig. 4a.

To better understand the weight dynamics of PEDOT:PSS-based complex synapses, a numerical calculation about the synaptic weight update of ideal PEDOT:PSS-based simple/complex synapses under repeated writing procedures has been conducted, as shown in Fig. 5. Fig. 5a displays the weight update of an ideal simple synapse under repeated potentiation/depression processes. The weight pattern can easily be repeated with the same writing procedure. Clearly, the potentiation part is symmetric to the depression part under ideal conditions. Fig. 5b displays the weight update of an ideal complex synapse with additional $C_2$ component under the same repeated potentiation/depression process. The weight pattern of the first writing cycle is different to that of the second cycle due to the existence of $u_2$. A periodic relation between $u_1$ and $u_2$ can be slowly reached as the number of writing cycles increases. The dynamic range of the weight of this complex synapse for a single potentiation (or depression) process after several potentiation/depression processes is smaller than that of a simple synapse, suggesting the "regularizing" effect from the coupled $C_2$ component. Furthermore, more coupled storage components can be added to the complex synapse ($C_{2\text{-}3}$ in Fig. 5c and $C_{2\text{-}4}$ in Fig. 5d). More writing cycles are needed as the coupled components increase to reach a weight dynamics pattern that better reproduces itself. This feature is more obvious in Fig. 5e and f, where more writing cycles are conducted and displayed. Also, the total weight change for a single potentiation (or depression) process decreases as the number of additional storage components increases. All these features extracted from the numerical calculations are consistent with what we have observed in the experiments.

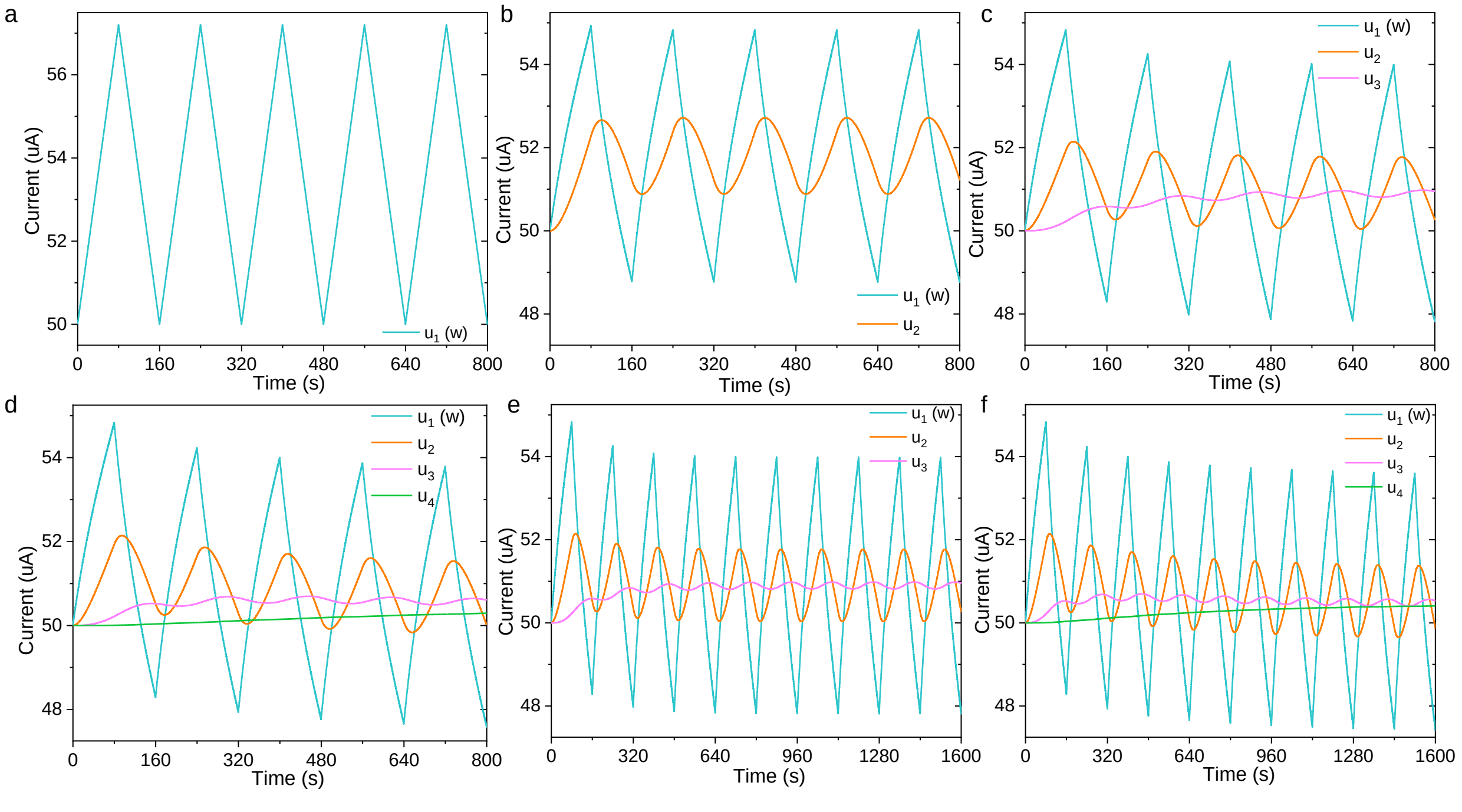

*Fig. 5* ***Numerical simulation of synaptic weight updates for ideal PEDOT:PSS-based simple/complex synapses under repeated writing procedures. a****, Synaptic weight update of a simple synapse (only $C_1$). Writing pulse: -/+1 V. Pulse frequency: 1 Hz. "on" state fraction: 50%. Pulse number for each potentiation/depression process is fixed to the same: 80.* ***b****, Synaptic weight update of a complex synapse (with additional $C_2$).* ***c****, Synaptic weight update of a complex synapse (with additional $C_2$ and $C_3$).* ***d****, Synaptic weight update of a complex synapse (with additional $C_2$, $C_3$ and $C_4$). The writing procedures in* ***b-d*** *are the same as that in* ***a****.* ***e****, Synaptic weight update of the same complex synapse in* ***c*** *but with more repeated writing cycles.* ***f****, Synaptic weight update of the same complex synapse in* ***d*** *but with more repeated writing cycles. The read voltage is set as 0.01 V for all these simulations.*

Finally, to demonstrate the computational and functional advantages of these features, neural network simulations for face familiarity detection utilizing PEDOT:PSS-based simple/complex synapses are conducted, as shown in Fig. 6 and Fig. S6-7. Fig. 6a displays the architecture of our face familiarity detection system[18]. A series of face images from VGGFace2[35, 36] are presented in sequence to the neural system through the input module. The input module contains a deep convolutional neural network, and its weights are frozen during online learning (pre-trained based on several face tasks, more information can be found in the Methods part). The memory module is the only part of the system containing plastic synapses which are continuously updated by the ongoing presentation of face patterns. Fig. 6b displays the task diagram of our face familiarity detection tasks. Two kinds of detection tests (familiarity detection (FD) and forced choice (FC)) and different working conditions (same pose (SP), different pose (DP) and random pattern (RP)) are adopted in our simulations. We focus on the recall performance of memories as a function of the number of other memories stored in the meantime, in other words, memory age. The quantitative metrics include ideal-observer signal-to-noise ratio (ioSNR), readout signal-to-noise ratio (rSNR), detection accuracy, and familiarity memory lifetime $t^*$ (defined as the age at which the test performance drops below some threshold). Three neural networks utilizing PEDOT:PSS-based simple synapses (number of memory module neurons N = 128, number of synapses = $128^2$, dynamical variable per synapse m = 1) with different learning rates (q = 1, 0.2, and 0.05, respectively) and an advanced neural network utilizing PEDOT:PSS-based complex synapses (N = $64^2$, m = 4, q = 1) are designed for the simulations. To make a fair comparison between PEDOT:PSS-based complex synapses and simple synapses, the total number of dynamic variables in these four neural systems is matched to be the same.

Fig. 6c-e show the results of familiarity detection accuracy and memory lifetime FD $t^*$ for FD tests with these four systems. As we can see, smaller learning rates for simple synapses can lead to a longer memory lifetime, at the expense of a lower initial detection accuracy and generalization ability. For the SP and RP cases, the familiarity memory lifetime of the neural network with PEDOT:PSS-based complex synapses is over 20 times longer than that of the neural network with simple synapses under the same learning rate, while for the DP case, the improvement factor is more than 4. Similar results can be obtained for FC tests (Fig. 6f-h). The neural network utilizing PEDOT:PSS-based complex synapses performs substantially better than the neural network utilizing simple synapse with equal learning rates (q = 1), not to mention that the former has a smaller number of neurons. Moreover, it still has an advantage compared to the neural network with a very small learning rate (q = 0.05) set for simple synapses to extend their memory lifetime. Consistent results of ioSNR and rSNR for these four simulated neural networks can be found in Fig. S6.

We can conclude from the face familiarity simulations that the memory model with PEDOT:PSS-based complex synapses prevails in memory capacity, and this advantage can become even more prominent in networks with a larger number of neurons due to the different scaling behaviors of simple and complex systems[1]. A larger number of dynamic variables for each complex synapse also helps to improve the performance of the complex neural networks (Fig. S7).

## Conclusions

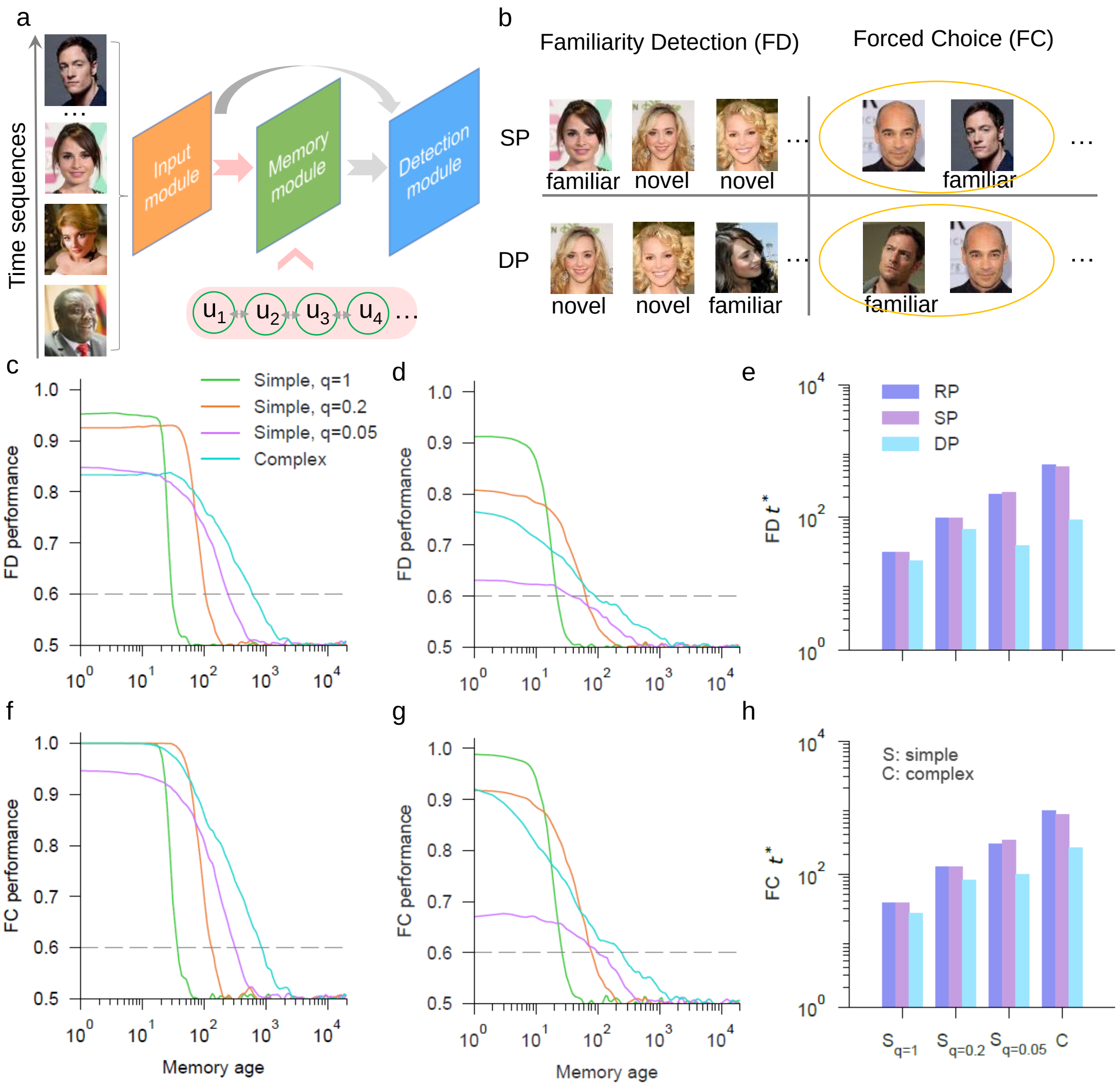

*Fig. 6* ***Neural network simulations for face similarity detection utilizing PEDOT:PSS-based simple/complex synapses. a****, Architecture of our face familiarity detection system. The neural network contains three modules: input module, memory module, and detection module. A series of face images are presented at a time in sequence to the neural system through the input module. The synapses between the input module and the memory module are plastic (indicated by pink arrow), while all other synapses are fixed after being pre-trained (indicated by grey arrows).* ***b****, Task diagram of face similarity detection. In a familiarity detection (FD) test, the network is required to determine whether a presented face image is familiar or novel. In a two-alternative forced choice (FC) test, two face images (exactly one familiar and one novel) are presented to the network, and the system is required to choose which one of the two is familiar. There are two working conditions for both tasks: same pose (SP) and different pose (DP).* ***c-e****, Comparison of the FD performance between four different neural networks: FD accuracy for same pose (****c****), FD accuracy for different pose (****d****), and familiarity memory lifetime FD t* for SP/DP/random-pattern (RP) cases (****e****). The memory lifetime t* is defined as the age at which the test performance drops below some threshold (0.6 for example, as indicated by the dashed gray line in* ***c*** *and* ***d****). Three neural networks contain PEDOT:PSS-based simple synapses ($N = 128^2$, $m = 1$) with different learning rates ($q = 1$, 0.2, and 0.05, respectively) and an advanced neural network contains PEDOT:PSS-based complex synapses ($N = 64^2$, $m = 4$, $q = 1$). The total number of synaptic dynamic variables for each neural network is the same. ).* ***f-h****, Comparison of the FC performance between four mentioned neural networks: FC accuracy for same pose (****f****), FC accuracy for different pose (****g****), and familiarity memory lifetime FC t* for SP/DP/random-pattern (RP) cases (****h****).*

In conclusion, by designing a PEDOT:PSS-based redox transistor together with a series of charge-diffusion-controlled storage components, we have realized a hardware implementation of the Benna-Fusi complex synapse. Effective and reliable controllability of weight updates of our PEDOT:PSS-based synapses via gate voltage pulses has been fully demonstrated. The multi-variable effect from coupled storage components has been confirmed by our experimental observations. Neural network simulations based on the hardware parameters of our complex synapse show a superior memory capacity during online learning for real-world applications compared to networks based on simple synapses. We believe that this hardware realization of Benna-Fusi complex synapses is a substantial step towards successful on-chip continual learning.

## Methods

**Preparation of PEDOT:PSS-based simple synapse/complex synapse and synaptic arrays.** PEDOT:PSS was purchased from Sigma-Aldrich (3.0-4.0% in $H_2O$). Firstly, interconnects for the polymer-based redox transistor were patterned on a silicon wafer with 300 nm thermal oxide using a standard photolithography method. 5/100 nm Ti/Au was deposited via e-beam evaporation to form interconnects after lift-off process. Secondly, the wafer with interconnects was coated with Parylene-C (~3 μm) as the shadow layer. No adhesion processing is needed for the coating of Parylene-C. Corresponding photolithographic patterning was performed on the coated wafer and oxygen-plasma etching (PlasMod O2 Asher) was utilized to define the transistor channel/gate area and areas for other storage components. The exposed areas for coupled storage components can be adjusted based on the previous design. Thirdly, PEDOT:PSS solution was prepared by adding 5 vol% ethylene glycol (Sigma Aldrich) and 1 vol% (3-glycidyloxypropyl)trimethoxysilane (Sigma Aldrich), the former of which can enhance the PEDOT:PSS morphology and the later one is a crosslinking agent to improve the mechanical stability. PEDOT:PSS solution was spin-coated on the wafer at 1000 RPM for 120 s and baked at 125 °C for 20 minutes. The Parylene-C shadow layer can be then peeled off with scotch tape to retain PEDOT:PSS only in the photolithographically well-defined redox-transistor channel/gate areas and areas for other storage components. Finally, the gel-state Nafion (as both the electrolyte and the diffusion media) was formed by drop-cast of solution (Nafion™ 117 containing solution, Sigma Aldritch) using a micropipette and then covering with deionized water. For the fabrication of PEDOT:PSS-based complex synapse array, a wafer with a diameter of 2 inches (MTI) was specially chosen.

**Electrical characterization of the potentiation/depression of PEDOT:PSS-based synapses.** The I-t curves with writing pulses were measured using Autolab PGSTAT302N potentiostat. The gating pulses for potentiation/depression were provided by an AFG1062 Tektronix Arbitrary Waveform Generator. The gating electrode (G) is connected to the source electrode (S) in series with a 50 MΩ resistance to reduce gate leakage of the device. All electrical characteristics of PEDOT:PSS-based synapses were measured under ambient conditions and room temperature.

**Numerical calculation of weight updates for PEDOT:PSS-based synapses.** The analytical solution for the system of partial differential equations involving Benna-Fusi model can be obtained with an eigen-decomposition method (more discussions can be found in the Supplementary Information). Codes in MATLAB based on the analytical solution were created. The weight updates of PEDOT:PSS-based synapses as a function of writing procedures can be calculated once all the parameters of system composition and initial conditions are given. In turn, certain parameters (e.g., coupling coefficient $g_{k,k+1}$) can be extracted by fitting if the weight updates as a function of writing procedure are provided.

**Neural network simulations for a face familiarity detection system utilizing PEDOT:PSS-based synapses.** We used face images from a public large-scale face data set VGGFace2 [35] as the stimuli fed into our system. Each image is resized and center-cropped to 224 × 224 pixels. Our system consists of the input module, the memory module, and the detection module. The input module uses a deep convolutional neural network (SE-ResNet-50), pre-trained on the MS-Celeb-1M data set and fine-tuned on the VGGFace2 data set, to extract the face feature vectors (2048 dimensions of the penultimate layer). The feature vectors are dimension-reduced to $N$ dimensions using principal component analysis, and then binarized with median values of each dimension ("input patterns" or "input neurons"). The memory module consists of $N$ memory neurons. The $i$-th memory neuron receives inputs from $j$-th input neurons ($j \neq i$), i.e., $y_i = sign(b_i + \sum_{j \neq i} w_{ij} x_j)$, where $w_{ij}$ and $b_i$ are the weights and biases, respectively. Weights and biases, corresponding to the $u_1$ variables, are updated by the Hebbian plasticity rule (i.e., $\Delta w_{ij} = x_i\, x_j$ and $\Delta b_i = x_i$). We set $C_1 = 1, C_2 = 1, C_3 = 2, C_4 = 4$ for our simulations. For simple synapses, we further controlled their learning rate (q), where the synaptic updates ($\Delta w_{ij}$ and $\Delta b_i$) are accepted with probability p and rejected with probability 1-p. Finally, the detection module computes the Hamming distance between the input pattern from the input module and the memory pattern from the memory module, to determine whether a face image is familiar or novel. For the mathematical details about the FD performance, FC performance, and signal-to-noise-ratio, see [18].

**Acknowledgment**

This work is supported by the US National Science Foundation under award number 1916652, and the NYSTAR Focus Center at Rensselaer Polytechnic Institute under award number C180117. M.K.B was supported by NIH (R01NS125298) and the Kavli Institute for Brain and Mind.

**Author contributions**

J.S., M.K.B. and L.Z. conceived and developed the idea. L.Z. prepared samples and performed electrical measurements. L.Z. and Y.H. processed the raw data. J.J. helped to design devices. R.L. did the numerical calculation for synaptic weight updates. J.A.L. and M.K.B. conducted neural network simulations for face familiarity detection. L.Z., J.J. and J.S. analyzed and interpreted the results. L.Z. prepared the initial manuscript draft. J.S., M.K.B. and J.A.L. revised it. All authors were involved in the discussion for data analysis and the writing of the manuscript. J.S. supervised the project.

**Additional information**

Supplemental materials: Analytical solution for Benna-Fusi model, Synaptic updates of a single-variable synapse under different conditions, Simulations for weight updates of PEDOT:PSS-based simple/complex synapses under different conditions, Fabrication of PEDOT:PSS-based complex synapse array, Neural network simulations for face detection utilizing PEDOT:PSS-based simple/complex synapses.

**Competing financial interests**
The authors declare no competing financial interests.

**Data availability**
Data are available upon reasonable requests from the corresponding authors.

**Code availability**
Codes are available upon reasonable requests from the corresponding authors.

# Supplementary Information

**L Zhang et al**

## Table of Contents

## Supplementary Discussions

### Analytical solution for the dynamical evolution of the Benna-Fusi model

1) We take four storage components as an example. According to the basic dynamical equation $C_k \frac{du_k}{dt} = g_{k-1,k}(u_{k-1} - u_k) + g_{k,k+1}(u_{k+1} - u_k)$, we have:

$$C_1 \frac{du_1}{dt} = g_{1,2}(u_2 - u_1)$$

$$C_2 \frac{du_2}{dt} = g_{1,2}(u_1 - u_2) + g_{2,3}(u_3 - u_2)$$

$$C_3 \frac{du_3}{dt} = g_{2,3}(u_2 - u_3) + g_{3,4}(u_4 - u_3)$$

$$C_4 \frac{du_4}{dt} = g_{3,4}(u_3 - u_4), \text{ SI (1)}$$

Equivalently, we have

$$C \frac{du}{dt} = Gu, \text{ SI (2)}$$

where $C = \begin{pmatrix} C_1 \\ C_2 \\ C_3 \\ C_4 \end{pmatrix}$, $u = \begin{pmatrix} u_1 \\ u_2 \\ u_3 \\ u_4 \end{pmatrix}$, $G = \begin{pmatrix} -g_{1,2} & g_{1,2} & 0 & 0 \\ g_{1,2} & -g_{1,2} - g_{2,3} & g_{2,3} & 0 \\ 0 & g_{2,3} & -g_{2,3} - g_{3,4} & g_{3,4} \\ 0 & 0 & g_{3,4} & -g_{3,4} \end{pmatrix}$, and also $u(0) = \begin{pmatrix} u_1^0 \\ u_2^0 \\ u_3^0 \\ u_4^0 \end{pmatrix}$.

If we set $v = C^{1/2}u$, then we have $\frac{dv}{dt} = C^{-1/2}GC^{-1/2}v$. We can perform the eigen-decomposition for the matrix $H = C^{-1/2}GC^{-1/2} = E\boldsymbol{\Lambda}E^T$.

Then we can get $v(t) = e^{(t-t_0)H}v(0) = E \begin{pmatrix} e^{(t-t_0)\lambda_1} & 0 & 0 & 0 \\ 0 & e^{(t-t_0)\lambda_2} & 0 & 0 \\ 0 & 0 & e^{(t-t_0)\lambda_3} & 0 \\ 0 & 0 & 0 & e^{(t-t_0)\lambda_4} \end{pmatrix} E^T v(t_0)$, SI (3)

Correspondingly, $u(t) = C^{-1/2}E \begin{pmatrix} e^{(t-t_0)\lambda_1} & 0 & 0 & 0 \\ 0 & e^{(t-t_0)\lambda_2} & 0 & 0 \\ 0 & 0 & e^{(t-t_0)\lambda_3} & 0 \\ 0 & 0 & 0 & e^{(t-t_0)\lambda_4} \end{pmatrix} E^T C^{1/2} u(t_0)$, SI (4)

2) In neural network training, we have potentiation/depression pulses to update the synaptic weight $u_1$, so we can modify equation SI (2) by introducing an input term:

$$C\frac{du}{dt} = Gu + \begin{pmatrix} s \\ 0 \\ 0 \\ 0 \end{pmatrix}, \text{ SI (5)}$$

where $s$ represents the weight change per unit time due to the imposed writing pulse on $C_1$.

Similarly, we have $\frac{dv}{dt} = C^{-1/2}GC^{-1/2}v + \begin{pmatrix} \frac{s}{\sqrt{C_1}} \\ 0 \\ 0 \\ 0 \end{pmatrix}$, SI (6). Again Performing the eigen-decomposition for

$H = C^{-1/2}GC^{-1/2} = E\boldsymbol{\Lambda}E^T$, we get $\frac{dE^Tv}{dt} = \boldsymbol{\Lambda}E^Tv + E^T\begin{pmatrix} \frac{s}{\sqrt{C_1}} \\ 0 \\ 0 \\ 0 \end{pmatrix}$, SI (7).

Let $\hat{v} = E^Tv$, $b = E^T\begin{pmatrix} \frac{s}{\sqrt{C_1}} \\ 0 \\ 0 \\ 0 \end{pmatrix}$, we have $\frac{d\hat{v}}{dt} = \boldsymbol{\Lambda}\hat{v} + b$, in other words, $\frac{d\hat{v_i}}{dt} = \lambda_i\hat{v_i} + b_i$, $i = 1,2,3,4$.

In terms  of these redefined variables, for an input $b$ that is constant after time $t_0$, we get $\hat{v_i}(t) = b_i(t - t_0) + \hat{v_i}(t_0)$, if $\lambda_i = 0$, SI (8) and $\hat{v_i}(t) = \left(\hat{v_i}(t_0) + \frac{b_i}{\lambda_i}\right)e^{\lambda_i(t-t_0)} - \frac{b_i}{\lambda_i}$, if $\lambda_i \neq 0$, SI (9), where $\hat{v_i}(t_0) = E^Tv(t_0) = E^TC^{1/2}u(t_0)$.

Then we obtain $u(t) = C^{-\frac{1}{2}}E\hat{v}(t)$, SI (10).

3) Considering the duration ratio $d$ $(0 \leq d \leq 1)$ for a single writing pulse in our experiments, we can further modify equation SI (5) into:

$$C\frac{du}{dt} = Gu + \begin{pmatrix} p \\ 0 \\ 0 \\ 0 \end{pmatrix}, \text{ where } p = \begin{cases} s, & t \in [nT_0, (n+d)T_0] \\ 0, & t \in ((n+d)T_0, (n+1)T_0] \end{cases}, \text{ SI (11)}$$

where $n$ is an integer. When $t \in [nT_0, (n+d)T_0]$, from SI (8-9), we first compute $\hat{v}(nT_0) = E^TC^{1/2}u(nT_0)$. Then we have:

$$u(t) = C^{-\frac{1}{2}}E\hat{v}(t), \text{ where } \hat{v}(t) = \begin{cases} b_i(t - nt_0) + \hat{v_i}(nT_0), & if\ \lambda_i = 0 \\ \left(\hat{v_i}(nT_0) + \frac{b_i}{\lambda_i}\right)e^{\lambda_i(t-nT_0)} - \frac{b_i}{\lambda_i}, & if\ \lambda_i \neq 0 \end{cases}, \text{ SI (12).}$$

When $t \in ((n+d)T_0, (n+1)T_0]$, from SI (4), we have:

$$u(t) = C^{-\frac{1}{2}} E \begin{pmatrix} e^{(t-(n+d)T_0)\lambda_1} & 0 & 0 & 0 \\ 0 & e^{(t-(n+d)T_0)\lambda_2} & 0 & 0 \\ 0 & 0 & e^{(t-(n+d)T_0)\lambda_3} & 0 \\ 0 & 0 & 0 & e^{(t-(n+d)T_0)\lambda_4} \end{pmatrix} E^T C^{\frac{1}{2}} u((n+d)T_0),$$ SI (13).

Since $\widehat{v_\iota}((n+d)T_0) = E^T C^{\frac{1}{2}} u((n+d)T_0)$, we obtain:

$$u(t) = C^{-\frac{1}{2}} E \begin{pmatrix} e^{(t-(n+d)T_0)\lambda_1} & 0 & 0 & 0 \\ 0 & e^{(t-(n+d)T_0)\lambda_2} & 0 & 0 \\ 0 & 0 & e^{(t-(n+d)T_0)\lambda_3} & 0 \\ 0 & 0 & 0 & e^{(t-(n+d)T_0)\lambda_4} \end{pmatrix} \widehat{v_\iota}((n+d)T_0),$$ SI (14), where

$\widehat{v_\iota}((n+d)T_0)$ can be calculated from SI (12).

## Supplementary Figures

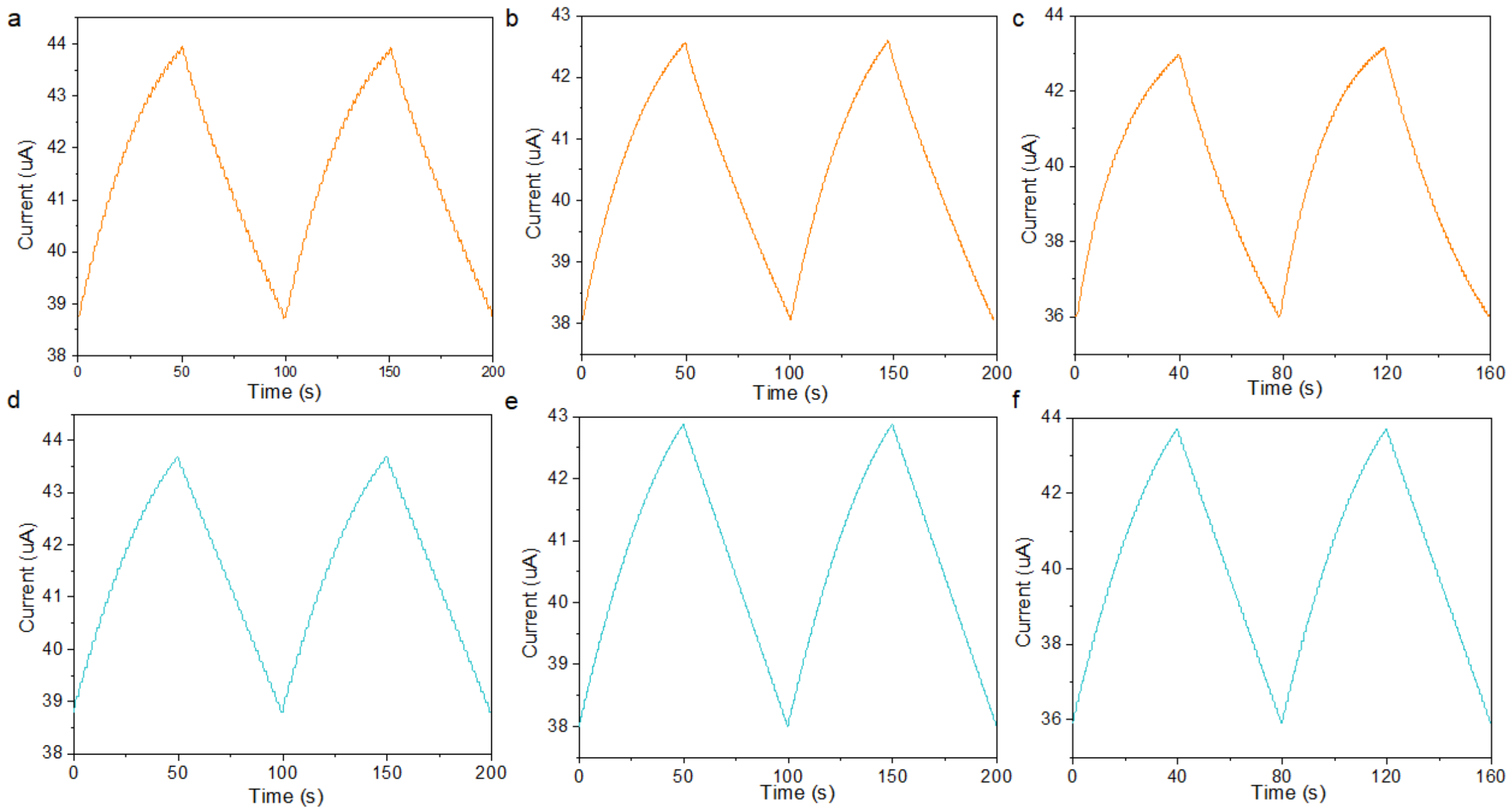

**Fig. S1 | Synaptic depression and potentiation of a single-variable synapse (only $C_1$) with writing pulses of different durations or of different strengths. a**, Pulse strength: -/+ 1 V. Pulse frequency: 0.5 Hz, "on" state percentile: 50%. **b**, Pulse strength: -/+ 1 V. Pulse frequency: 1 Hz, "on" state percentile: 50%. **c**, Pulse strength: -/+ 2 V. Pulse frequency: 1 Hz, on state percentile: 50%. **d**, Numerical simulation of **a**. **e**, Numerical simulation of **b**. **f**, Numerical simulation of **c**. The magnitude of *s* in the simulation is fixed to be the same in **d** and **e**, and in **f** it is twice as large as in **e**. Note that *s* is not a constant but a weak function of $u_1$.

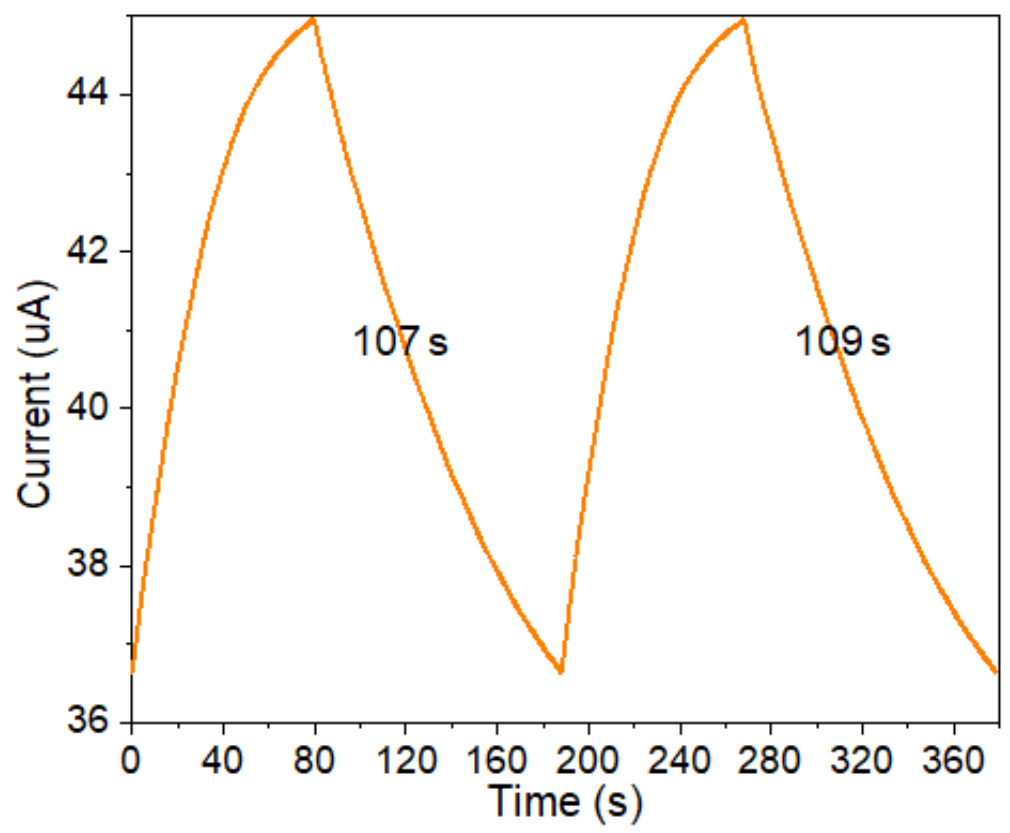

**Fig. S2 | Weight update of a single-variable synapse (only $C_1$) near extrema.** There are 80 writing pulses for potentiation of single-variable $C_1$ while it takes about 108 pulses for depression to its original state. *s* becomes smaller and smaller as the current increases to a high level for the potentiation process. On the contrary, *s* is large when the current level is high for the depression process, and it becomes smaller and smaller as the current decreases to a very low level. *s* is also dependent on the state of the storage component $C_1$ ($u_1$), besides the gate voltage, especially when $u_1$ has a large dynamic range.

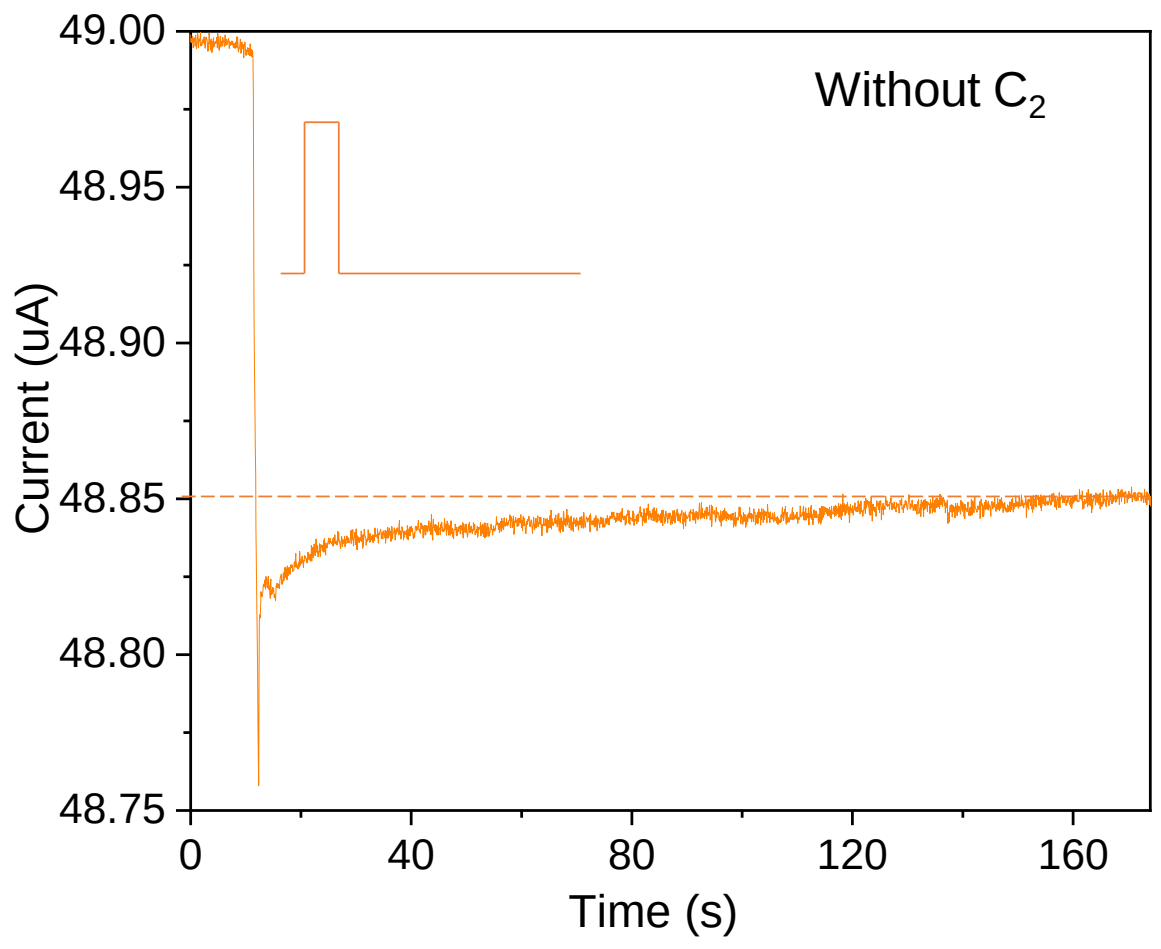

**Fig. S3 | The weight change of a simple synapse as a function of time after a large writing pulse (+3 V, 1 s duration).** A voltage of 0.01 V is used for the read operation.

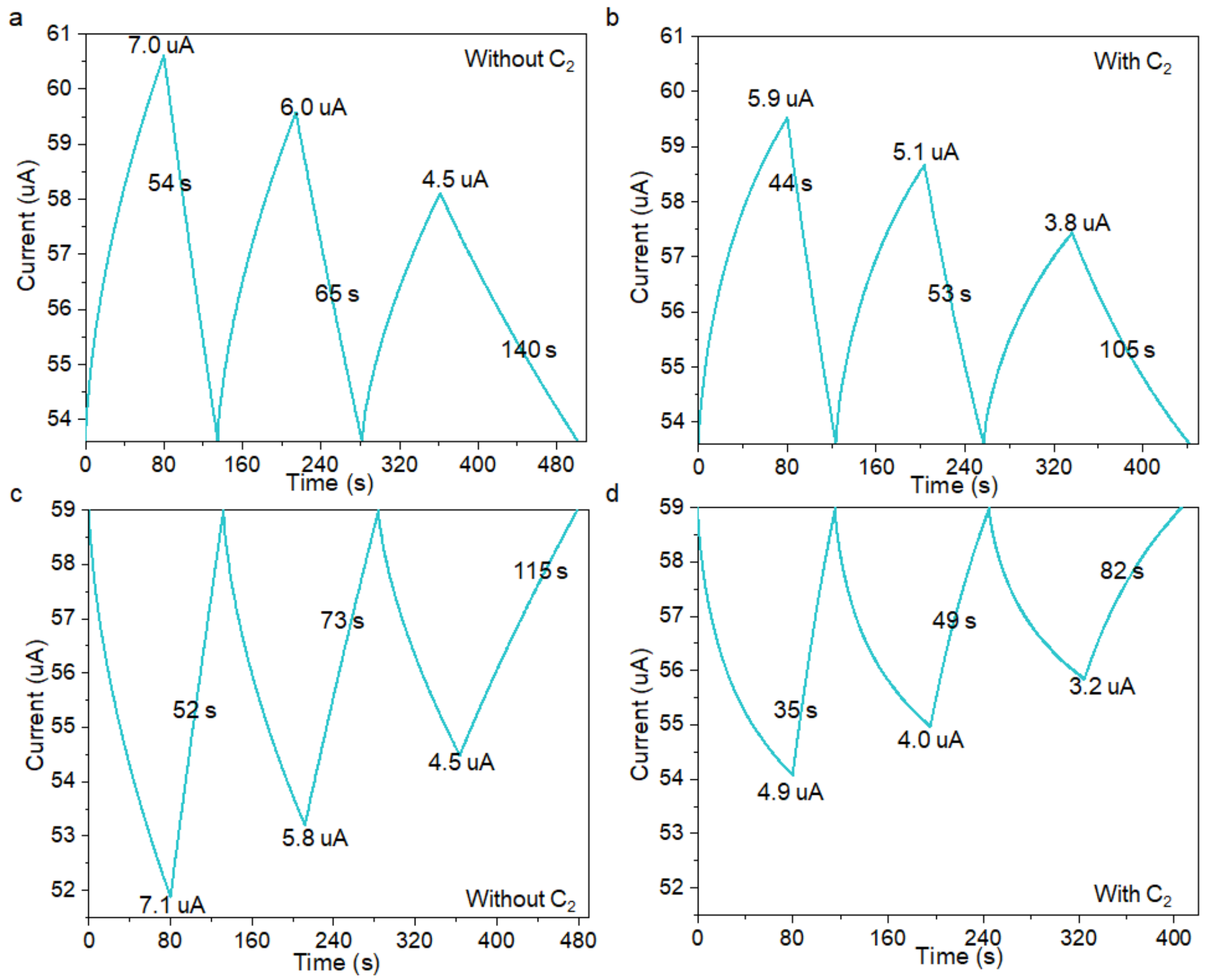

**Fig. S4 | Simulations for weight updates of PEDOT:PSS-based simple/complex synapses with writing pulses of different "on" state percentile. a**, Simulation of synaptic potentiation and depression of a simple synapse (only $C_1$). Pulse strength: -/+1 V. Pulse frequency: 1 Hz. "on" state percentile from left to right; 66.6%, 50%, 33.3%. Pulse number for each potentiation process is the same: 80. **b**, Simulation of synaptic potentiation and depression of the same synapse in **a** but with the additional $C_2$ connected. **c**, Simulation f synaptic depression and potentiation of a simple synapse (only $C_1$). Pulse strength: -/+1 V. Pulse frequency: 1 Hz. "on" state percentile from left to right; 66.6%, 50%, 33.3%. Pulse number for each potentiation process is the same: 80. **d**, Simulation of synaptic depression and potentiation of the same synapse in **c** but with the additional $C_2$ variable connected. A coupling coefficient of $g_{12}/C_1 \cong 2^{-8}$ for these complex synapses (with additional $C_2$ storage component) can be extracted from the calculated results.

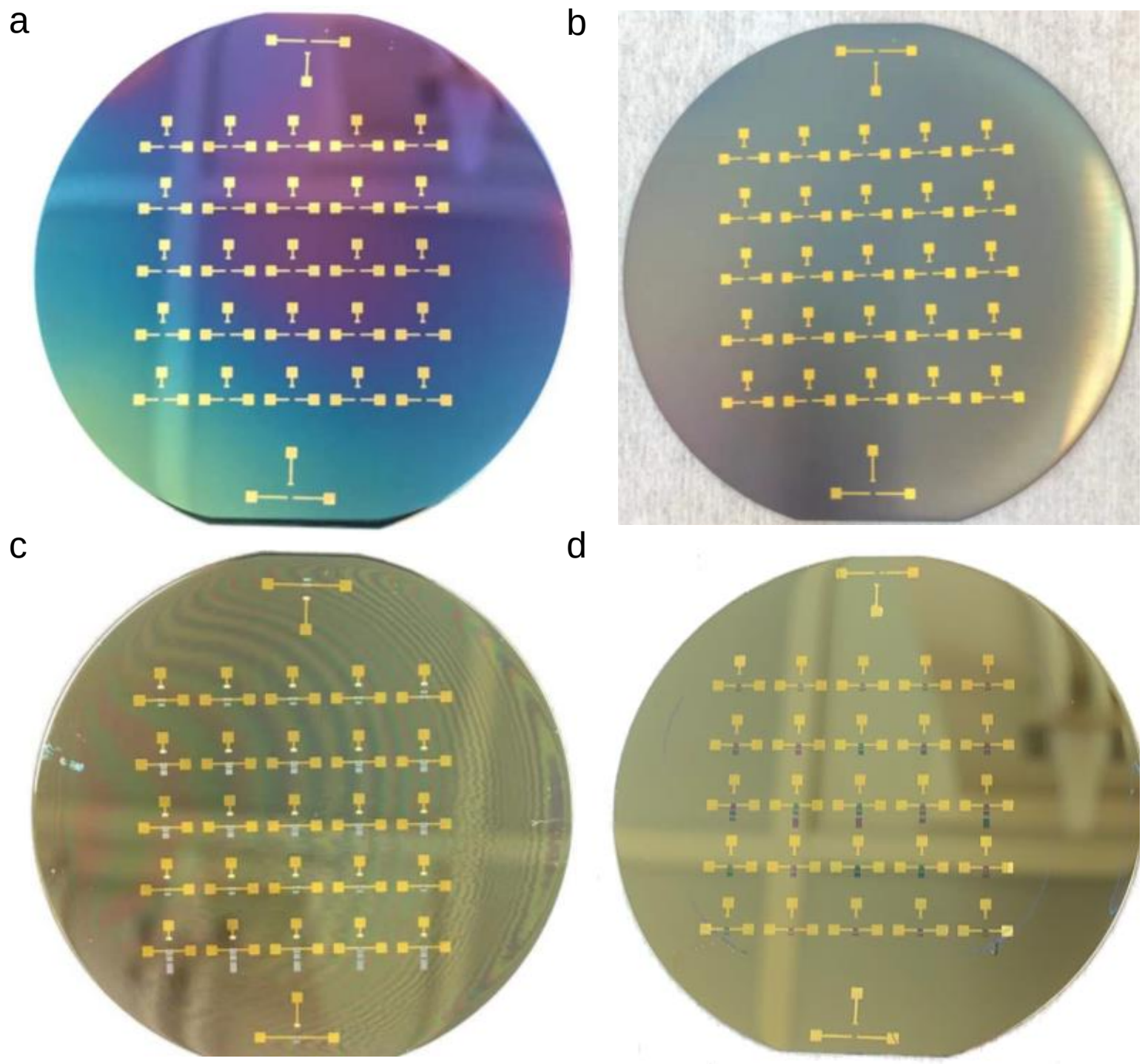

**Fig. S5 | Fabrication of PEDOT:PSS-based complex synapse array. a**, 5/100 nm Ti/Au interconnects patterned on a two-inch silicon wafer with 300 nm thermal oxide using a standard photolithography method. **b**, Thin Parylene-C film deposited as the shadow layer. **c**, Oxygen-plasma etching is used after photolithographic patterning to define the transistor channel/gate area and areas for other storage components. **d**, Parylene-C shadow layer can be then peeled off with scotch tape after PEDOT:PSS coating and heating to retain PEDOT:PSS only in the photolithographically well-defined redox-transistor channel/gate areas and areas for other storage components.

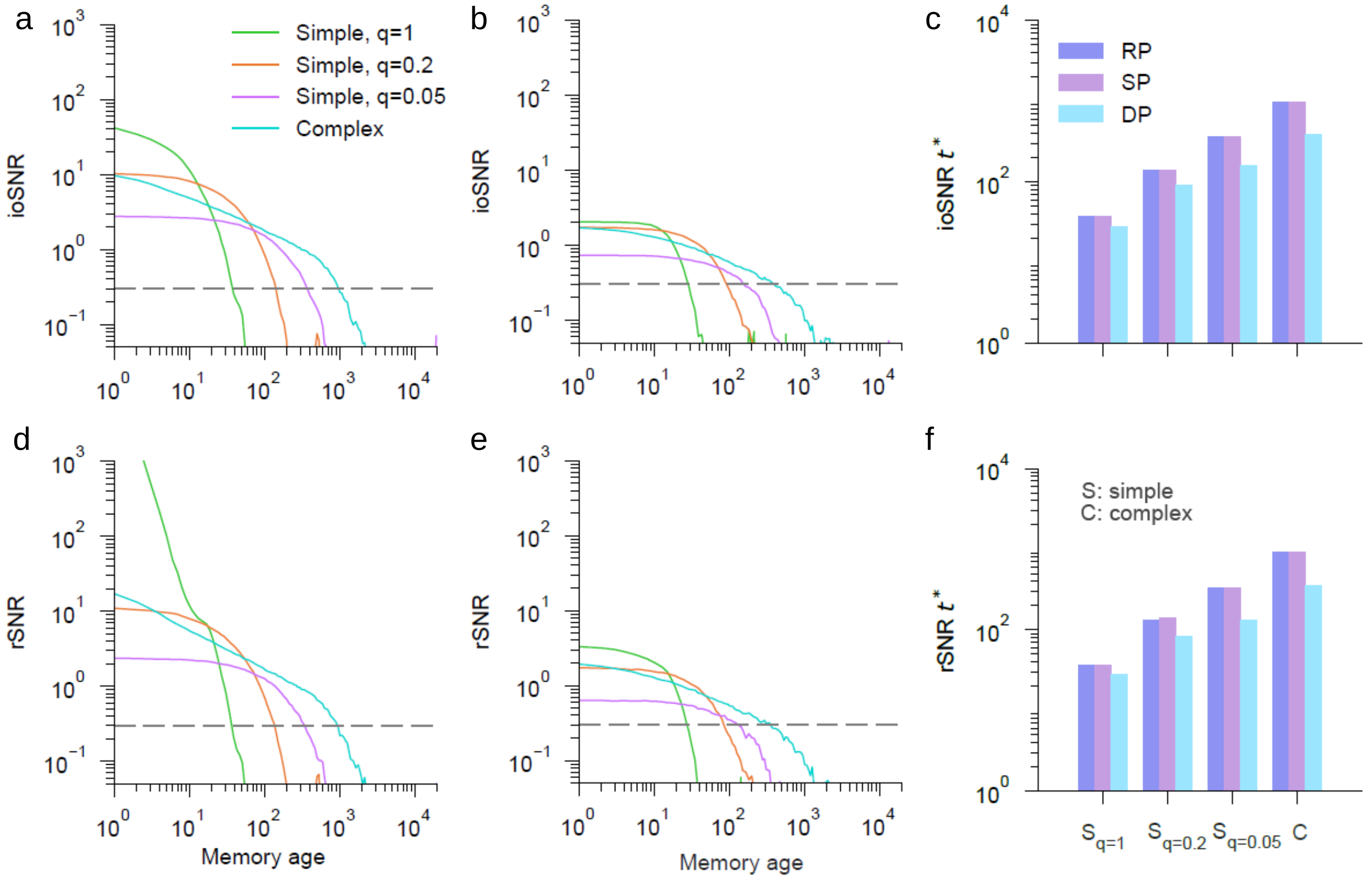

**Fig. S6 | Neural network simulations for face similarity detection utilizing PEDOT:PSS-based simple/complex synapses.** Three neural networks utilizing PEDOT:PSS-based simple synapses ($N = 128^2$, $m = 1$) with different learning rates ($q$ = 1, 0.2, and 0.05, respectively) and an advanced neural network utilizing PEDOT:PSS-based complex synapses ($N = 64^2$, $m = 4$, $q = 1$) are designed. **a-c**, Comparison of the ioSNR performance between four different neural networks: ioSNR for same pose (**a**), ioSNR for different pose (**b**), and familiarity memory lifetime ioSRN t* for SP/DP/RP cases (**c**). The memory lifetime t* is defined as the age at which the SNR drops below some threshold (0.3 for example, as indicated by the dashed gray line in **a** and **b**). The total number of synaptic dynamic variables for each neural network is the same. **d-f**, Comparison of the rSNR performance between the four above-mentioned neural networks: rSNR for same pose (**d**), rSNR for different pose (**e**), and familiarity memory lifetime rSNR t* for SP/DP/RP cases (**f**).

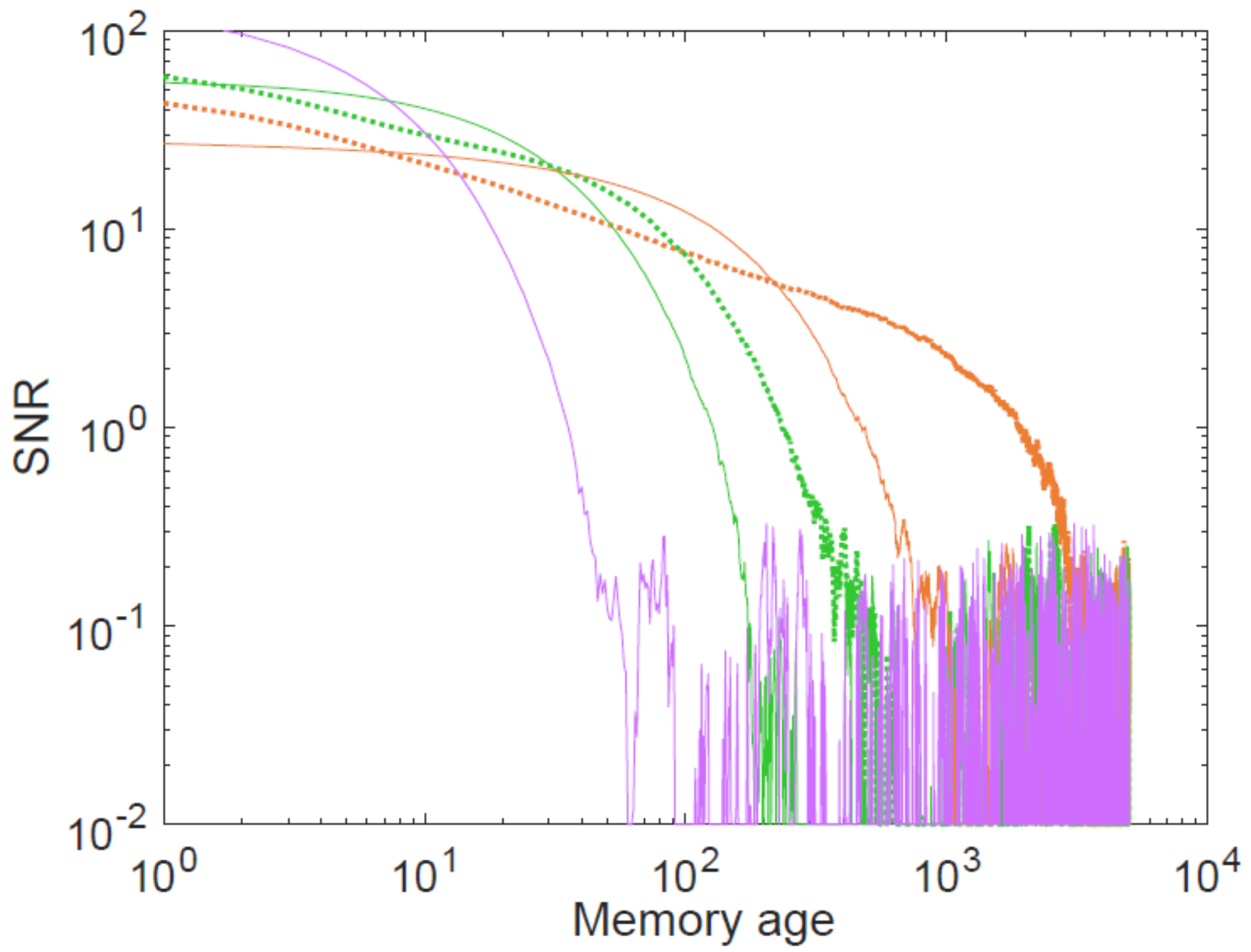

**Fig. S7 | Neural network simulations utilizing PEDOT:PSS-based simple synapses and complex synapses with different numbers of dynamical variables.** The dashed lines represent our experimental models with two dynamic variables per synapse ($C_1 : C_2 = 1 : 1$, green) and four dynamic variables per synapse ($C_1 : C_2 : C_3 : C_4 = 1 : 1 : 2 : 4$, orange). The purple solid curve shows the simulation results for simple synapses with $m = 1$. They have a large initial SNR, which decays quickly (exponentially). For the green and orange solid curves, we simulate simple synapses whose decay timescales are set to be equal to those of the complex models with $m = 2$ and $m = 4$, respectively. This reduces the initial SNR but extends the memory lifetime (slower exponential decay) compared to the purple curve. However, they cannot obtain the same memory lifetimes as the corresponding more complex models.